# Role Engine Implementation for a Continuous and Collaborative Multi-Robot System

Behzad Akbari, *Member, IEEE,* Zikai Wang, Haibin Zhu, *Senior Member, IEEE,* Lucas Wan, *Student Member, IEEE*, Ryan Adderson, *Student Member, IEEE*, and Ya-Jun Pan, *Senior Member, IEEE*

*Abstract* — **In situations involving teams of diverse robots, assigning appropriate roles to each robot and evaluating their performance is crucial. These roles define the specific characteristics of a robot within a given context. The stream actions exhibited by a robot based on its assigned role are referred to as the process role. Our research addresses the depiction of process roles using a multivariate probabilistic function. The main aim of this study is to develop a role engine for collaborative multi-robot systems and optimize the behavior of the robots. The role engine is designed to assign suitable roles to each robot, generate approximately optimal process roles, update them on time, and identify instances of robot malfunction or trigger replanning when necessary. The environment considered is dynamic, involving obstacles and other agents. The role engine operates hybrid, with central initiation and decentralized action, and assigns unlabeled roles to agents. We employ the Gaussian Process (GP) inference method to optimize process roles based on local constraints and constraints related to other agents. Furthermore, we propose an innovative approach that utilizes the environment's skeleton to address initialization and feasibility evaluation challenges. We successfully demonstrated the proposed approach's feasibility, and efficiency through simulation studies and real-world experiments involving diverse mobile robots.**

## I. INTRODUCTION

Multi-robot teams have found applications in various domains, such as surveillance [1], inspection [2], rescue operations [3], automation [4], and logistics [5]. However, the collaboration among these agents can pose significant challenges, especially when the robots vary in terms of hardware, size, and functionalities within dynamic environments. To adapt to the environment, robots need to dynamically adjust their behavior and optimize their task sequences to minimize energy consumption and prevent collisions. The process role is a term used to describe the sequence of actions carried out by a robot according to its assigned role. This concept is derived from the E-CARGO model and the Role-Based Collaboration (RBC) theory, as outlined in references [6], [7]. Enabling collaboration and management interaction based on process roles can significantly impact this approach. For example, in a firefighter scenario, a robot may be designated as an extinguisher corresponding to a destination point $x_N(t)$ and process role includes determining the optimal trajectory, motion dynamics, and activation timing and angle for extinguishing fires (Fig. 1). Similarly, an aerial robot may assume the role of a mapper, responsible for relocating to the best position to provide information or the fire's location to other agents. The trajectory of these multivariant actions (process role) shapes the behavior of the robot according to the role it fulfills.

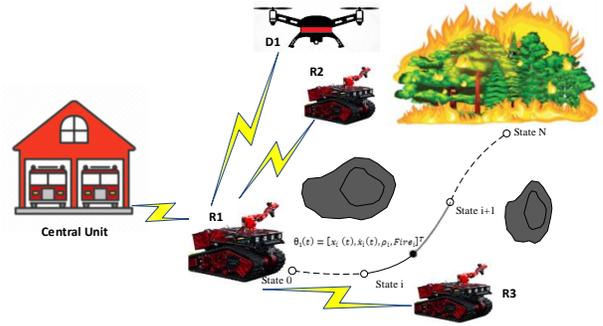

**Fig. 1.** The process role multivariant function of a ground Firefighter robot(R1) involves a sequence of actions $\theta_i(t)$, which include the dynamic characteristics $(x_i(t), \dot{x}_i(t))$ and the functional aspects of the extinguisher, such as the angle at which it operates and the command flag for extinguishing fires $(\rho_i, Fire_i)$.

An ideal role engine incorporates mechanisms to assess the feasibility of assigned roles [6] [7]. It is capable of assigning suitable roles and adapting them as needed. The role engine can handle teams of different sizes, ranging from small groups to large-scale deployments. It mostly employs a hybrid control approach, combining centralized control for role assignment and initialization (role negotiation, agent evaluation, and role assignment) and decentralized control for task execution (role playing). This allows for effective coordination among robots while maintaining flexibility and independence. Optimization techniques are used to improve robot behavior and performance. The role engine also includes mechanisms to detect robot malfunctions and failures, triggering replanning

Manuscript received ****, 2021. This work was supported in part by the Natural Sciences and Engineering Research Council, Canada (NSERC) under Grant RGPIN-2018-04818 and the Innovation for Defence Excellence and Security (IDEaS) program from the Canadian Department of National Defence (DND) under grant CFPMN2-051. Any opinions and conclusions in this work are strictly those of the authors and do not reflect the views, positions, or policies of – and are not endorsed by – IDEaS, DND, or the Government of Canada.

B. Akbari, Z. Wang and H. Zhu are with Collaborative Systems Laboratory (CoSys Lab), Department of Computer Science and Mathematics, Nipissing University, North Bay, Ontario, Canada (Email: behzada@nipissingu.ca, zwang529@my.nipissingu.ca, haibinz@nipissingu.ca)

L. Wan, R. Adderson and Y.-J. Pan are with the Advanced Control and Mechatronics Laboratory (ACM Lab), Department of Mechanical Engineering, Dalhousie University, Halifax, Nova Scotia, Canada (Email: lucas.wan@dal.ca, ryan.adderson@dal.ca, yajun.pan@dal.ca).

processes to reassign roles or adjust strategies for the team to maintain functionality and complete tasks [6]. The primary focus of this paper was on the adaptation of roles and the optimization of process roles. By considering the process role as a continuous multivariant probabilistic function, we can optimize its posterior density based on the factors occurring in the scenario. Gaussian process (GP) inference, a probabilistic method, allows us to make predictions and perform probabilistic reasoning on unknown functions. The combination of GP inference and factor graphs for application in motion planning scenarios is referred to as Gaussian Process Motion Planning (GPMP), as introduced in [8] [9]. This research introduced the concept of utilizing the GP function as a process role that can be influenced by factors or dynamic events within the system. While finding the optimal process roles in multi-agent systems is a computationally challenging task, GP inference can provide an approximately optimal solution by efficiently incorporating kinematic and size constraints in a continuous domain. The approximation of optimality is achieved through the Gaussian assumption and an iterative method for solving nonlinear optimization [10, 11, 12]. In the case of multi-agent scenarios, GP inference can be extended to incorporate new shared factors, such as robot collision, in the optimization problem [13].

This paper presents an advanced role engine design that encompasses a wide range of desirable characteristics. These characteristics are achieved by employing a flexible hybrid approach, which ensures excellent scalability in various settings with different numbers of robots and environmental conditions. The approach closely mimics real-world situations by using centralized control during training and initialization stages, and decentralized control during operational phases.

The paper utilizes Group Role Assignment (GRA) [6] [14] as a central mechanism to assign initial process roles, reducing costs associated with each source-destination pair. In decentralized role playing, GP inference is implemented to establish consensus among autonomous robots, effectively leveraging shared information during role-playing. A novel technique is proposed where a simplified representation of the environment is used to create a dedicated map for each type of robot.

Furthermore, the paper improves GP inference to generate multiple continuous process roles within dynamic environments. The role engine proposed in this study addresses the challenge of managing a diverse group of robots with unknown roles and arbitrary initial positions. It initializes process roles centrally based on robot size, evaluates feasibility, generates a qualification matrix, and determines the most suitable assignments. After identifying the initial process roles to handle dynamic factors and uncertainty introduced by different autonomous agents, each robot is capable of autonomous movement while staying connected to others to achieve consensus in their process role. They create and navigate along an approximately optimal continuous trajectory that avoids collisions and adheres to their individual limitations. The main contributions of this paper can be summarized as follows:

1. The first development of a comprehensive role engine for collaborative multi-robot systems, encompassing role

initialization, association, assignment, and role-playing using the RBC methodology.
2. Modeling a multivariant probabilistic function for the process role using GP inference, allowing for optimization based on local or shared factors.
3. Practical implementation of the system for multi-robot pathfinders.

## II. BACKGROUND AND PRELIMINARIES

### A. Role-Based Collaboration (RBC)

The RBC methodology employs roles as its primary mechanism to facilitate collaboration [15] [16]. The fundamental abstract model of RBC is E-CARGO (Environments - Classes, Agents, Roles, Groups, and Objects) [17] [18]. An RBC-based role engine encompasses role negotiation, agent evaluation, GRA, and role-playing [19]. Role negotiation involves clarifying and verifying roles to ensure successful collaboration among agents in a given scenario. If a viable solution exists, agents can be assessed for each role, and a qualification matrix can be generated based on the process roles and agents involved. The role engine handles dynamic role assignment. However, the existing RBC role engine does not address process role optimization. This paper introduces an enhanced RBC model, which includes a component for role adaptation and employs multivariate GP inference to model the process role. This enables the optimization of the process role by considering environmental factors and information from other agents.

### B. Gaussian Process Inference

A GP is a probabilistic model or probability distribution that describes a collection of functions [20]. It represents smooth, continuous-time functions using a limited number of states. GP interpolation can be efficiently computed using a gradient-based optimization algorithm. The Markov model is a mathematical framework used to express GP [21]. In GP inference, a Bayesian approach is employed, incorporating prior constraints and the likelihood of events. The Maximum A Posteriori (MAP) estimation is utilized to find an approximately optimal solution. The prior constraint promotes smoothness, and encourages start and end points, while the likelihood function ensures collision-free paths with reduced energy consumption [9]. , as shown in:

$$\theta^* = \underset{\theta}{\text{argmax}} \ P(\theta|e_1, e_2, .., e_{N_f-1}), \qquad (1)$$

where $\theta$ is the multivariant vector-valued function representing the agent behavior (process role), and $e_1, e_2, .., e_{N_f-1}$ are a set of random binary events of interest. The posterior distribution of $\theta$ given $e_1, e_2, .., e_{N_f-1}$ can be derived from the prior and likelihood by Bayes rule:

$$P\left(\theta|e_1, .., e_{N_f-1}\right) \propto \ P(\theta)L(e_1|\theta)..L\left(e_{N_f-1}|\theta\right) \qquad (2)$$

Similar to [11] this rule is represented as the product of a series of factors solvable with a factor graph,

$$P(\theta|e) \propto \prod_{n_f=1}^{N_f} f_{n_f}\left(\theta_{n_f}\right), \qquad (3)$$

where $f_{n_f}$ are factors on state subsets $\theta_{n_f}$. It is shown in [11] that this MAP problem can be solved efficiently in linear time by exploiting the sparsity.

### B-1. GP Markov model

Based on [21] [11], a linear time-varying stochastic differential equation (LTV-SDE) form of GP inference can be written as:

$$\dot{\theta}(t) = A(t)\theta(t) + u(t) + F(t)w(t), \tag{4}$$

where $u(t)$ is the known system control input, $A(t)$ and $F(t)$ are time-varying matrices of the system, and $w(t)$ is generated by a white noise process. The white noise process is itself a zero-mean GP:

$$w(t) \sim GP\big(0, Q_C \delta(t - t')\big), \tag{5}$$

where $Q_C$ is the power-spectral density matrix and $\delta(t - t')$ is the Dirac delta function. The solution to the initial value problem of this LTV-SDE is in the form of mean and covariance [21]:

$$\tilde{\mu}(t) = \Phi(t, t_0)\mu_0 + \int_{t_0}^{t} \Phi(t, s)u(s)ds, \tag{6}$$

$$\tilde{K}(t, t') = \Phi(t, t_0)K_0\Phi(t', t_0)^T + \int_{t_0}^{\min(t, t')} \Phi(t, s)F(s)Q_C F(s)^T \Phi(t', s)^T ds, \tag{7}$$

where $\Phi$ is the state transition matrix and $\mu_0, K_0$ are the mean and covariance, respectively, at $t_0$. The Markov property of (4) results in the sparsity of the inverse kernel matrix $K^{-1}$ which allows for fast inference [21]. Then the proportion of GP prior, can be written as follows:

$$P(\theta) \propto \exp\left\{-\frac{1}{2} \parallel \theta - \mu \parallel_K^2\right\}. \tag{8}$$

where $\parallel . \parallel_K^2$ is the Mahalanobis distance with covariance K.

### B-2. The collision avoidance likelihood function

In GP inference, constraints are formulated as rules that the trajectory must obey. For example, the likelihood function of collision avoidance indicates the probability of being free from collisions with other robots or obstacles. All likelihood functions are defined as a distribution in the exponential family for a single robot, given by:

$$L_s(\theta_i; e_i) \propto \exp\left\{-\frac{1}{2} \parallel h(\theta_i) \parallel_{\Sigma_{abs_i}}^2\right\}, \tag{9}$$

where $h(\theta_i)$ is a Hinge loss function for a given current configuration $\theta_i$, $e_i$ is the corresponding event, such as no collision, and $\Sigma_{abs_i}$ is the hyperparameters of distribution [22]. For a multi-agent case, a new factor for collision between robots needs to be combined as in [13]:

$$L_m(\theta_i; e_i) \propto$$
$$\exp\left\{-\frac{1}{2} \parallel h(\theta_i) \parallel_{\Sigma_{abs_i}}^2\right\} \prod_{\substack{l=1 \\ i' \neq i}}^{m} \exp\left\{-\frac{1}{2} \parallel g(\theta_i, \theta_{i'}) \parallel_{\Sigma_{mul}}^2\right\}, \tag{10}$$

where $g(\theta_i, \theta_i')$ is a vector-valued function that defines the cost of two agents $i$ and $i'$ being close to each other, $m$ is the number of agents and $\Sigma_{mul}$ is a hyperparameter of the distribution. The specific likelihood and obstacle cost function used in this paper is slightly different from (10), detailed in Section III-C (Role playing).

### B-3. MAP inference

By solely focusing on obstacle detection and collision avoidance, we can formulate the MAP posterior estimation using equations (2), (8), and (9) as follows:

$$\theta_j^* = \underset{\theta}{\operatorname{argmin}} \{ \frac{1}{2} \parallel \theta_j - \mu_j \parallel_K^2 + \frac{1}{2} \parallel h(\theta_j) + h1(\theta_j, \theta_{0..n}) \parallel_{\Sigma_{obs}}^2 \}, \tag{11}$$

which is a non-linear least square problem that can be solved by iterative algorithms such as Gauss-Newton or Levenberg-Marquardt until convergence. The corresponding linear equation is as follows:

$$\theta^* = \underset{r}{\operatorname{argmin}} \parallel A\theta - b \parallel^2, \tag{12}$$

where $A \in \mathbb{R}^{n_f \times n_s}$ is the measurement Jacobian consisting of $n_f$ measurement rows and $b$ is an $n_s$-dimensional vector computable similar to iSAM2 [23].

## III. ROLE ENGINE FORMULATION

This study introduces the implementation of a role engine specifically designed for a diverse group of robots operating in an environment filled with obstacles. These robots have different hardware, sizes, and capabilities, and they freely navigate in a dynamic environment. The role engine utilizes a hybrid approach that combines role negotiation and assignment through a central system (Algorithm 1) to initiate robots for role-playing. Although the role-playing itself is decentralized, the robots establish connections with each other and exchange information to reach a consensus using a distributed channel, as depicted in Fig. 2. In our approach, each robot's role is determined based on its destination and process role, which includes trajectory functions such as dynamics and functionality. The destination positions are not predefined and can involve various formations, such as circles, squares, or lines. By leveraging Fig. 2, our role engine initializes, assigns, and optimizes the best process roles for the robots through the central computer to minimize overall cost, with a particular focus on factors like smoothness and obstacle avoidance.

To assess the suitability of each robot for a given destination, we employ trajectory cost as a measure in agent evaluation. The trajectories and outcomes of the GRA algorithm serve as an initial guideline for robots to engage in role-playing. During the role-phase, the robots take into account both the initial process roles derived from GRA and newly observed factors such as collision likelihood, to optimize process roles. This enables the robots to adapt their actions and roles in response to a dynamic environment.

### A. Role negotiation and feasibility check

This subsection is dedicated to role negotiation; relevant information is extracted from the environment, scenario, and available resources to assess the feasibility of the problem. It is essential to thoroughly investigate all potential solutions in order to assign appropriate roles to the agents. An action field representation called the Environment Map (E-Map) is constructed to streamline the search process and expedite evaluation, incorporating weighted graphs. The E-Map needs to be tailored to suit different types of robots. For instance, certain obstacles may be disregarded for aerial vehicles, while large robots may have restricted access to certain paths. Based on the robots' parameters, an E-Map is generated from the original environment image, which can be sourced from a camera capturing real-life situations. The Role-Negotiation algorithm examines all feasible paths and generates possible initial paths. If there are no viable initial paths connecting the source and destination, the problem is deemed infeasible. Algorithm 2 provides a detailed explanation of the Role-Negotiation algorithm. The algorithm's output includes environment representations based on the type of robot, signed distance fields (as discussed in Section IV.B), and initial process roles. Further information regarding the E-Map and Auxiliary nodes can be found in Appendix A.

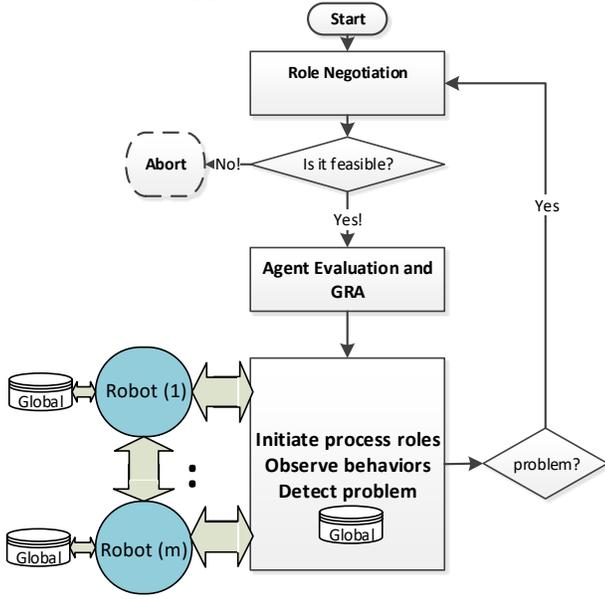

**Fig. 2.** The general flowchart of the proposed algorithm.

---

**Algorithm 1**: Central Computer Main Function
---
**Input:** $\mathcal{A}, \mathcal{R}, e, arg$ // Agents, roles (Destinations ID), env, and arguments

1: *Feasibility*, $Env_\tau$, $Sdf_\tau$, *Init_paths* ← Role-Negotiation($\mathcal{A}, \mathcal{R}, e, arg$)
2: **if** *Feasibility* ==False:
3:    Abort()
4: **endif**
5: $\lambda = arg.\lambda$  //coefficent weight for different costs
6: $\bar{\theta}$ ← GRA($Env_\tau$, $Sdf_\tau$, *Init_paths*, $\lambda$) // Optimal initial paths
7: Publish($\bar{\theta}$, *Shared_Channel*) //set the distributed storage
7: **for** $j$ **in** range($\bar{\theta}$. *Length*): //number of process roles
8:    $a_j$ ← Define_agent($\bar{\theta}$,j, $Env_{\tau(j)}$)
9: **endfor**
10: **for** $j$ **in** range($\bar{\theta}$. *Length*):
11:    RolePlaying.start($a_j$) //Start role playing
12: **endfor**
13: **while** true:
14:    $\bar{\theta}$, NewMap = subscrib(*Shared_Channel*)
15:    **if** Prob($\bar{\theta}$, NewMap) == true: //problem detected
16:       **if** Not_Negotiable:
17:          Abort()
18:       **Else**
19:          $arg$ = UpdateArg($\bar{\theta}$) //update argumnets
20:          **Go**(Line1)    //Renew the negotiation
21:       **endif**
22:    **endif**
23: **endwhile**

The original GP inference (GPMP algorithm) [11] makes the assumption of a direct path from the source to the destination, represented by the red dashed line in Fig. 3. However, this simplistic approach is overly optimistic in complex environments with intricate arrangements of obstacles. To ensure practical and realistic results, the algorithm in [11] requires offline parameter adjustment. On the other hand, in [24], a fully supervised method is employed to determine the GP parameters. While GP is typically a semi-supervised method that can be trained using diverse training environments, our approach does not prioritize that aspect. Instead, we incorporate E-Maps, which are customized maps designed for each robot type. E-Maps facilitate the identification of a quicker shortest path and auxiliary nodes. In our approach, the initial path is established as a collection of line segments connecting these auxiliary nodes, which serves as an approximation of the shortest route from the source to the destination. The distribution of steps along these connected line segments, as illustrated in Fig. 3, determines the initial path for each robot. This incorporation of E-Maps allows us to achieve a faster GRA and a narrower range of parameter tuning compared to the conventional GPMP method. For additional in-depth details, please refer to Appendix A. The computational complexity of the Role-Negotiation stage is denoted as $O(mn\mathfrak{e})$, where $m$ and $n$ represent the number of robots and roles, respectively, and $\mathfrak{e}$ represents the number of edges in the E-Map graph (as explained in Appendix A). The complexity analysis indicates that Lines 7-15 have a complexity of $O(mn)$, while Line 9 has an additional complexity of $O(\mathfrak{e})$.

---

**Algorithm 2**: Role-Negotiation Algorithm
---
**Input:** $\mathcal{A}, \mathcal{R}, e$ //Agents, roles (Destinations), environment
**Output:** *Feasibility*, $Env_\tau$, $Sdf_\tau$, *Init_paths*

1: *Init_paths* ← [] // Initialize to an empty list
2: **for** $\tau$ **in** $\mathcal{A}.\Gamma$:    // For all robot types, initiate environments
3:    $Env_\tau$ ← Environment($e, \tau$) // Dedicated for each robot's type
4:    $Sdf_\tau$ ← SignedDistanceField($Env_\tau$)
5:    $E\_Map_\tau$ ← Make_Env_map($e, \tau$)
6: **endfor**
7: **for** $a$ **in** $\mathcal{A}$:    // for all agents(sources)
8:    **for** $r$ **in** $\mathcal{R}$:  // for all roles (destinations)
9:       $Aux\_noods[a,r]$ ← Find_Aux($E\_Map_{\tau(a)}, a, r$)
10:       $Init\_paths[a,r]$ ← Make_Init_Path($Aux\_noods[a,r]$, $a,r$)
11:    **Endfor**
12:    **if**{Union of all possible paths to $r$ }=={∅}:
13:       *Feasiblity*=False
14:    **endif**
15: **endfor**
16: **return** *Feasibility*, $Env_\tau$, $Sdf_\tau$, *Init_paths*

### B.  GRA and agent evaluation

Suppose that the system is workable (minimum feasible assignment found for all roles), and based on role negotiation, the qualification of each agent can be computed for a process role and used to construct the $Q$ matrix [6] [14] [25]. The qualification of each agent in that role is defined as a floating-point value result of the process role cost function. Using the generated $Q$ matrix and the GRA algorithm, the optimal process role assignment can be obtained from the team's viewpoint. The $Q$ matrix in the proposed problem presents the cost for each robot's process role.

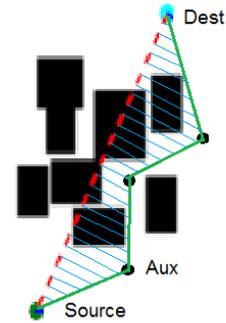

**Fig. 3.** Initial path based on E-Map (green path) compared to GPMP(red dash-line)

*Qualification matrix:* For every trajectory $\theta$, the qualification value is calculated using generic cost functions. Similar to GP inference, the objective function of each trajectory is defined by combining a function, $F_{gp}$, derived from the Gaussian process prior, which penalizes deviations from the prior mean to maintain smooth trajectories and a conflict cost function that considers the costs of obstacles and collisions associated with other robots, aiming to avoid such collisions:

$$q[\theta] = \lambda F_{gp}[\theta] + F_{conf}[\theta] , \qquad (13)$$

where the balance between the two functionals is regulated by the parameter $\lambda$, where the prior functional $F_{gp}$ is generated through the GP formulation.

$$F_{gp}[\theta] = \tfrac{1}{2} \parallel \theta - \mu \parallel_{\mathcal{K}}^2, \qquad (14)$$

---

**Algorithm 3:** GRA and Process role initialization

**Input:** $Env_\tau$, $Sdf_\tau$, $Init\_paths$, $\lambda$
**Output:** OptPaths   //Initial and optimized process roles

1:   **for** $a$ in $\mathcal{A}$:
2:     **for** $r$ in $\mathcal{R}$:   //for all agents and roles
3:       *$Init \leftarrow Init\_path[a,r]$*
4:       *if $Init == \emptyset$:* //if no path exists
5:         $Q[a,r] = Inf$
6:       **else** //if the initial path exists
7:         $\hat{\theta}_{a,r} = Init$ //Init path
8:         $ii = 0$ // Number of iterations
9:         **while** true:   //converging loop
10:           $d\theta_{a,r} = GP.Inference(Init, a, r, Env_{\tau(a)}, Sdf_{\tau(a)})$
11:           $\hat{\theta}_{a,r} = \hat{\theta}_{a,r} + d\theta_{a,r}$ //Update
12:           $\mathrm{Err}_{a,r} = F_{conf}(\hat{\theta}_{a,r}) + \lambda F_{gp}(\hat{\theta}_{a,r})$
13:           $ii = ii + 1$
14:           **if** $Convergence(d\theta_{a,r}, ii, \mathrm{Err}_{a,r})$ or ii > IterN:
15:             $Q[a,r] = \mathrm{Err}_{a,r}$ //qualification matrix
16:             *Break*
17:           **endif**
18:         **endwhile**
19:       **endif**
20:     **endfor**
21:   **endfor**
22:   $T_r$, $T_c$ =GRA(Q)   //Find the best assignment vectors
23:   **for** $i$ in $T_r$:   //for $i$ in row vector $T_r$
24:     OptPaths[$i$]$= \hat{\theta}_{T_r(i),T_c(i)}$
25:   **endfor**
26:   **return** OptPaths

---

A conflict cost function $F_{conf}$, is the combination of collision with obstacles and other agents. Similar to the approach used in [22] for sphere-shaped robots, the conflict function computes the arc-length parameterized line integral of the workspace obstacle and collision cost:

$$F_{conf}[\theta] = \int_{t_0}^{t_N} \int_\beta c(x) \parallel \tfrac{d}{dt} x \parallel dt , \qquad (15)$$

where $x$ is forward kinematic that maps robot configuration to workspace and $c(.)$ is the workspace cost function that penalizes the set of body points $\beta$ when they are around an obstacle and coming from the Signed Distance Field (SDF) of the environment [22]. The derivative, $\tfrac{d}{dt}x$ is the velocity of a body's center point in the workspace. Based on [11], multiplying the norm of the velocity with the cost in the trajectory integral above gives an arc-length parameterization. Given the $Q$ matrix, the GRA will find assignment matrix $T$, summarized in vectors $T_r$, $T_c$ using the Hungarian algorithm [14] (Table II). Regarding the complexity of Algorithm 3, taking into account the sparsity of GP inference [10], the complexity remains $O(mn\,N)$, within the nested loop. Here, $m$ represents the number of agents, $n$ represents the number of roles, and $N$ represents the length of the trajectory state. The overall complexity for process role initialization and GRA is $O(mn\mathrm{e} + mn\,N + m^3)$ due to the utilization of the Hungarian algorithm in GRA (e represents the number of edges in the E-Map graph).

### C. Role-playing

Our algorithm supports an interactive environment where autonomous robots are initialized and begin their actions by the central computer. To gather information about the environment and the robots' positions, we utilize a central camera that captures a top-view image of the environment. Alternatively, future implementations can obtain this information from an aerial mapper. Consensus in our role-playing occurs through the sharing of process roles between agents using message passing and a shared channel. This sharing of process roles is crucial for avoiding collisions between robots and addressing dynamic situations. A common approach to prevent inter-robot collisions is to treat nearby robots as stationary objects and employ Hinge losses similar to other obstacles. Another method, as described in [12, 13], introduces a new factor to penalize the proximity of agents. In our proposed method, we take a slightly different approach compared to [12, 13]. We examine conflicts among all the process roles at each time step and generate a dedicated collision potential field matrix for the agent. Fig. 5 provides an example to illustrate this concept.

**Table II:** Assignment matrix $T$ for the 4-robot scenario.

| $T$ | $r_1$ | $r_2$ | $r_3$ | $r_4$ |
|-----|-------|-------|-------|-------|
| $a_1$ | 0 | 0 | 1 | 0 |
| $a_2$ | 1 | 0 | 0 | 0 |
| $a_3$ | 0 | 1 | 0 | 0 |
| $a_4$ | 0 | 0 | 0 | 1 |

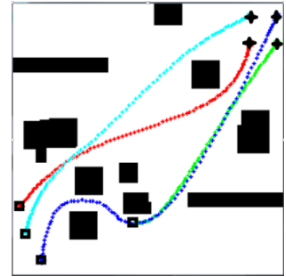

**Fig. 4.** The initial process roles were obtained through GRA for a scenario involving four agents navigating to unlabeled star points.

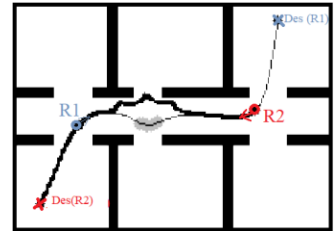

**Fig. 5.** Moving two robots in a narrow hallway and computing conflict field (the gray area) in timestep $k$ for robot R2(thicker path).

The role-playing process is detailed in Algorithm 4, where the prediction state at step $k$ is determined based on previous states, the robot's position, and collision likelihoods. Similar to the GP inference [11], the states are updated using an iterative approach. To address environmental changes and conflicts among agents, a function called *MakeConflictField(..)* has been created. This function compares process roles, updates the environment, and enforces penalties in cases of conflicts. Additionally, a straightforward color tracking algorithm is employed to track the robot's position. The calculated process role will get published in the shared channel. Considering the limited number of states for process roles, the complexity of Algorithm 4 at each time step primarily depends on GP inference, which has a linear relationship with the number of states $N$ [10]. Additionally, capturing conflicts between process roles has a complexity of $O(n^2)$ for n roles overall, resulting in a complexity of $O(N + n^2)$.

---

**Algorithm 4:** Role-playing Agent

**Input:** $Env_t$, $Init\_path$, $j$, $arg$ //Dedicated map and initial path for the agent $j$

1: **for** $k$ in total_time_step.range :
2:     $\bar{\theta}_k$, $NewEnv_j \leftarrow Subscribe(Shared\_Channel)$
3:     $NewSdf_j \leftarrow MakeConflictField(\bar{\theta}_k, NewEnv_j, k)$
4:     $Pos_j \leftarrow TrackPosition(NewEnv_j, k)$
5:     $d\theta_j \leftarrow GP.Inference(\bar{\theta}_j, Pos_j, NewEnv_j, NewSdf_j)$//Update state
6:     $\bar{\theta}_j = \bar{\theta}_j + d\theta_j$
7:     Publish ($\bar{\theta}_j$, $Shared\_Channel$)
8: **endfor**

---

## IV. IMPLEMENTATION DETAILS

In this part, more implementation details about process role optimization will be covered. To maintain simplicity, we focus on treating the process role as identical to the trajectory without taking into account any additional behavioral functions. The Levenberg-Marquardt iterative algorithm is used to solve the non-linear least-squares optimization problem in Eq. (11). The optimization is stopped if a maximum iteration is reached or if the relative decrease in error is smaller than a threshold.

### A. GP prior:

A constant-velocity prior is used for modeling the GP, including the positions and velocities of robots. For a robot $j$ the state includes:

$$\theta_j(t) = \begin{bmatrix} x_j(t), & \dot{x}_j(t) \end{bmatrix}^T, \tag{16}$$

where $\theta_j(t)$ is the position and velocity of the $j$-th robot in the group of $m$ robots. This model is a white-noise acceleration model

$$\ddot{x} = w(t). \tag{17}$$

The Markovian model in Eq. (4) is given by:

$$A(t) = \begin{bmatrix} 0 & I \\ 0 & 0 \end{bmatrix}, u(t) = 0, F(t) = \begin{bmatrix} 0 \\ I \end{bmatrix} \tag{18}$$

For $\Delta t_k = t_{k+1} - t_k$, the state transition matrix and process noise in Eqs. (6) and (7) are:

$$\Phi(t,s) = \begin{bmatrix} I & (t-s)I \\ 0 & I \end{bmatrix}, Q_{k,k+1} = \begin{bmatrix} \frac{1}{3}\Delta t_k^3 Q_c & \frac{1}{2}\Delta t_k^2 Q_c \\ \frac{1}{2}\Delta t_k^2 Q_c & \Delta t_k Q_c \end{bmatrix}. \tag{19}$$

The time propagation, denoted as $t - s$, involves the use of the identity matrix $I$ and the power-spectral density matrix $Q_c$ derived from the temporal kernel. The constant velocity GP prior is centered on a trajectory with zero acceleration. By incorporating this prior, the approach aims to reduce actuator acceleration in the configuration space. This interpretation provides a physical understanding of smoothness within this framework.

### B. Collision-avoidance likelihood:

An SDF matrix [22] is used for finding the collision likelihood, as in Fig. 6. The problem of finding the minimum signed distance from the robot surface to any obstacles is converted to see the difference between the signed distance of the sphere center and the sphere radius to any obstacle. An obstacle cost function for any trajectory $\theta_i$ is then completed by computing the Hinge loss:

$$h(\theta_i) = c(x(\theta_i, S)), \tag{20}$$

where $x$ is the forward kinematics for sphere $S$, $c$ is the Hinge loss function:

$$c(z) = \begin{cases} -d(z) + \epsilon & if\ d(z) \leq \epsilon \\ 0 & if\ d(z) > \epsilon \end{cases}, \tag{21}$$

where $d(z)$ is the signed distance from any point $z$ in the workspace to the closest obstacle surface, and $\epsilon$ is a 'safety distance' indicating the boundary of the 'unsafe area' near obstacle surfaces. The signed distance $d(z)$ is calculated from an SDF matrix. In this particular application, the SDF is specific to the type of robot and its interaction with other robots. The covariance parameter in Eq. (9) is defined as $\Sigma_{obs_i} = \sigma_{obs} I$.

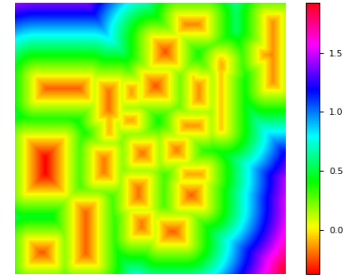

**Fig. 6.** SDF specific to a provided map containing obstacles

### C. Motion constraints:

Motion constraints exist in real-world robot team planning and formation problems and should be considered during trajectory estimation. For example, constraints may exist on initial and destination states as well as other states along the trajectory. These constraints are treated as prior knowledge on the trajectory states with very small uncertainties. Constraints can model such a penalty function $f(\theta_c)$, where $\theta_c$ is the set of states involved and can be incorporated into a likelihood function,

$$L_{constraint}(\theta) \propto \exp\left\{-\frac{1}{2} \| f(\theta_c) \|_{\Sigma_c}^2\right\}, \tag{22}$$

where $\Sigma_c = \sigma_c I$ and $\sigma_c$ is an arbitrary variance for this constraint, indicating how 'tight' the constraint is. For example, for speed limit constraint, the penalty function would be a difference between the current velocity and maximum velocity.

## V. EXPERIMENTAL STUDIES

A GPMP2 Python library [11] has been enrolled for process role optimization and OpenCV for image processing and simulations. The SciPy Optimize library is employed in Python to solve the optimization problem in GRA. All experiments are performed on a laptop with 8 Intel Core i7-4910MQ @ 2.90GHz CPUs, 16GB RAM. The approach to 2D path planning problems is tested with complex environment distributions and different robot types with size constraints.

### A. Simulation tests

#### 1- Obstacle distribution and feasibility test

In this scenario, four homogeneous robots are considered in six different environments, shown in Fig. 7, with varying obstacles and robot sizes (small, medium, large, and different combinations). The destination points are organized in formations resembling blue circles. Comparisons are conducted between the proposed algorithm and the traditional GPMP. The sensitivity of both algorithms to parameter settings using E-Map and without E-Map are examined for GRA. Considering parameters ($\sigma_{obs} = 0.05, 0.10, 0.15, 0.2$), the algorithms aim to find the approximate optimal paths from the sources to destinations using either the normal GPMP with a straight line as the initial path or the initial path from the E-Map. The percentage of feasibility, average error, and average converging time in both cases are measured for six different environments and four sets of parameters for a total of 24 scenarios.

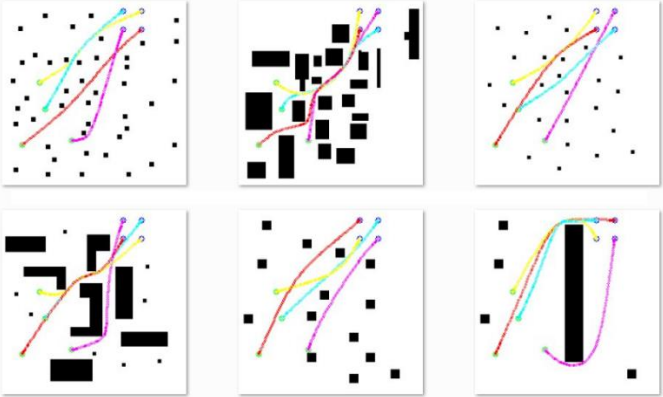

**Fig. 7.** The first experiment of the proposed algorithm with E-map for $\sigma_{obs} = 0.15$ (the green circles are the sources, and the blue circles are the destinations)

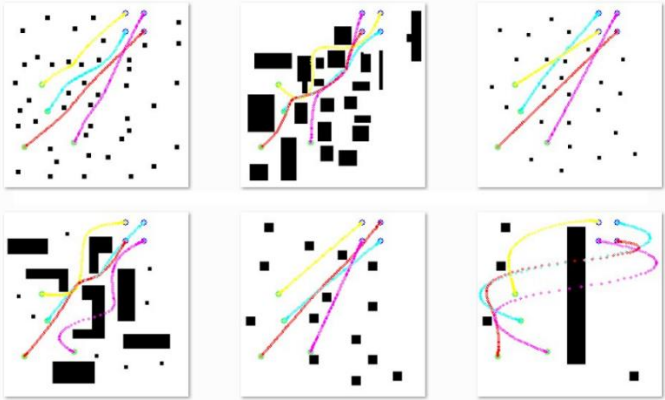

**Fig. 8.** GPMP without E-map for $\sigma_{obs} = 0.15$ , (the green circles are the sources, and the blue circles are the destinations)

The performance criteria of the two algorithms are presented in Table III. The feasibility is shown in terms of the average of success or failure. In the case of failures, there is at least a point on the graph where the center of the robot touches the obstacle. The average converging time is the average number of iterations to converge in each algorithm. Note that only the converging times for the feasible cases in GPMP are considered.

**Table III:** Comparing the results of two algorithms

| | Feasibility | Overall cost | Converging time |
|---|---|---|---|
| Our method | 100% | 0.00052 | 3.31 |
| GPMP | 57.14% | 0.01346 | 7.49 |

## 2- Testing of GRA advantages

In the second set of simulations, six robots are considered in two groups of different sizes. Three small sizes (radius = 0.02 m) and three large size robots (radius = 0.1 m). Similar to the previous experiment, 24 scenarios are tested for four ranges of parameters ( $\sigma_{obs} = 0.05, 0.10, 0.15, 0.2$ ) and six different environments. The environments are shown in Fig. 9, where the space between the obstacles is not necessarily larger than the robot size. An E-Map is used for both cases. In the first test, roles are associated with each robot base on the Nearest Neighbor (NN) algorithm means we associate the closest destination to sources. In the second algorithm, the GRA is used to test feasibility and role negotiation based on the cost of the robots, similar to Eq. (13). In Table IV, the feasibility, cost, and converging time for both algorithms are compared.

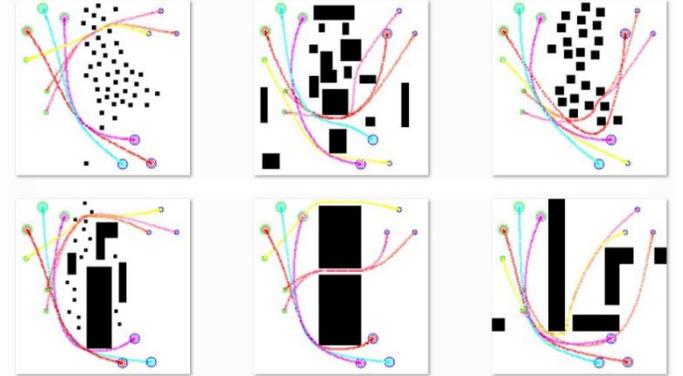

**Fig. 9.** The proposed algorithm under GRA for $\sigma_{obs} = 0.15$, (green circles with different radius are the robots with different sizes in the source, and blue ones are the destinations)

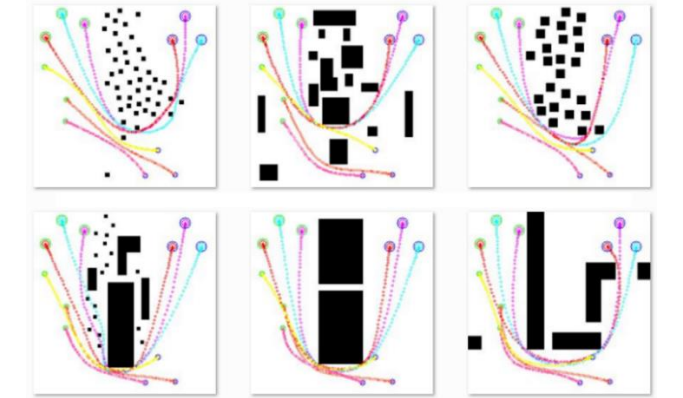

**Fig. 10.** GPMP using NN algorithm for $\sigma_{obs} = 0.15$, (the green circles with different radius are the robots with different sizes at the sources, and blue circles are the destinations)

**Table IV:** Comparing the algorithms for RBC model

| | Feasibility | Overall cost | Converging time |
|---|---|---|---|
| GRA | 100% | 0.000869 | 3.11 |
| NN | 100% | 0.001749 | 3.53 |

As shown in Table IV, the feasibility in both cases is 100% because of the E-Map. The proposed GRA algorithm significantly improves the average cost and converging time.

## 3-Role and task-based role-playing

We implemented a simulation case for two collision avoidance scenarios using a task or role-based collaboration. We considered four robots traveling from arbitrary sources to unlabeled destinations in a $5 \times 5\ m$ environment, including obstacles. In task-based collaboration, in Fig. 11A, robots share only their last positions and compute their distance to find shared collision likelihood, and in role-based collaboration, Fig.10B, they use all the process roles and compute their time-trajectory distances. In Fig. 11, the advantage of collaborating with the process role instead of tasks is shown with improved smoothness and reliability costs. We used the average jerk (change of acceleration) of the trajectories as a smoothness cost. The smoothness factor improved from 0.006 to 0.0003, and the minimum distance between all the robots also improved from 0.18 m to 0.23 m.

*4-Narrow hallway comparison with MA-GPMP [13]*

In this simulation, we compared our algorithm against the MA-GPMP [13] in a narrow hallway scenario. In Eq. (10) in the MA-GPMP algorithm, a shared factor for the inter-robot distance is extended to the optimization problem for each point. In our algorithm, we compute a cost function for each process role that conflicts with other process roles. The goal is to switch the positions of two robots that start in two rooms separated by a narrow hallway. The width of the hallway is smaller than the combined diameter of the two robots, meaning that robots cannot pass each other in the hallway. The size of the environment and settings are set to be similar to those in [13]. The result can be seen in Fig. 12. In our method, we penalize process roles for conflict areas and try to minimize the conflict by updating process roles close to the place where the conflict is happening. This is compared to the MA-GPMP method that tries to avoid the collision at the beginning (Fig. 12B). From these results, we can conclude that our approach makes more efficient use of the available space and is better suited for constrained environments than MA-GPMP.

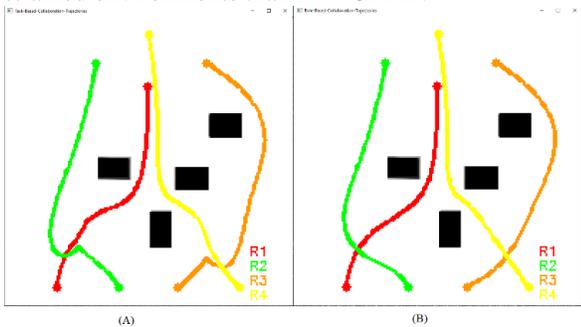

**Fig. 11.** The comparison of task-based (A) and role-based (B) role-playing.

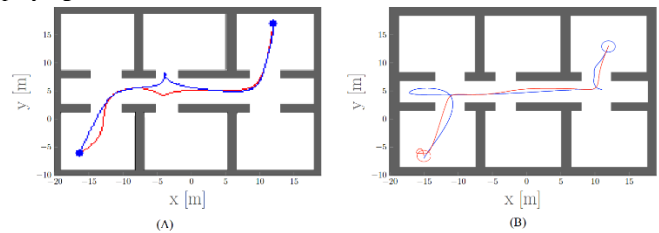

**Fig. 12.** The comparison of our proposed algorithm (A) against MA-GPMP algorithm (B).

### B. Experimental on the real robot

Experimental studies are carried out with four mobile robots on an indoor floor with a size of $5 \times 5$ m , filled with different size black obstacles (black boxes), and one camera installed to capture the scene. Raw frames, the E-Map, and the SDF matrix, are available in the resolution of 200 by 200 for each dedicated group of robots at any time. As shown in Fig. 13, two types of mobile robots with different sizes are used: a) two Pioneer 3-DX mobile robots with a length of 0.33 m between the two-wheel centers, effective linear velocity range of 0.1-0.7 m/s, and angular velocity range of 0.25-1.5 rad/s; b) two TurtleBot3 Burger mobile robots with the length of 0.16 m between the two-wheel centers, a maximum linear velocity of 0.22 m/s, and maximum angular velocity of 2.84 rad/s. As our proposed process role model only uses positions and linear velocity information about the agents, a separate controller based on [26] is used to develop a kinematic model of a two-wheel mobile robot and generate inputs of a linear velocity and an angular velocity. The algorithms are implemented firstly in Gazebo in Robot Operating System (ROS) and then are deployed to real-world robots. An example of the agents operating in a Gazebo environment is shown in Fig. 14.

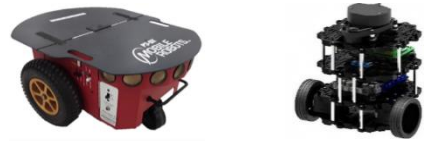

(a)Pioneer 3DX mobile robot      (b) TurtleBot3 Burger

**Fig. 13.** Experimental Setup at ACM Lab, Dalhousie University

In this test, four robots of two different sizes are commanded to form a diamond in a static environment. This test is first conducted in a simulated Gazebo environment and then applied to real robots. The robots communicate through a shared channel network in the Advanced Control and Mechatronics (ACM) lab at Dalhousie University. The experimental environment is shown in Fig. 16. The approximately optimal process roles generate 100 points. The main algorithm runs in the central computer and can efficiently assign robots with the proper roles in the environment with narrow and wide passages to minimize the costs. The trajectories of the robots in the experiments are shown in Fig. 15.

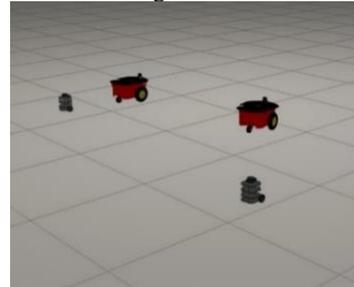

**Fig. 14.** An example of two Pioneer 3DX robots and two Turtlebot3 robots in a simulated Gazebo environment.

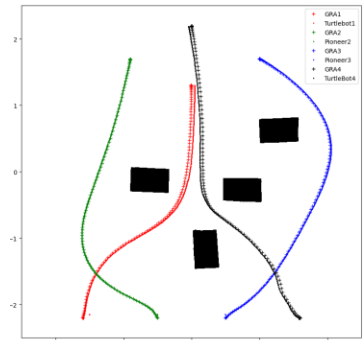

**Fig. 15.** Experimental results of the commanded (+) and actual trajectories (.) of the four robots.

The experimental results, as shown in Figs. 15 and 16, demonstrate the initial positioning of robots based on their unique characteristics and size. This arrangement enables efficient navigation through obstacles by leveraging the central computer. Once the autonomous robots begin their actions, they establish communication through the shared ROS channel and utilize the top view camera's image to perceive the environment. To monitor the robot positions, a color-based tracking algorithm is employed. The process roles are updated at each step as the robots progress toward their destinations. It's worth noting that a small error is observed in Figure 15, which can be attributed to tracking issues resulting from low resolution. Future work could address this by incorporating additional sensors to enhance the accuracy of robot localization in the environment.

## VI. Conclusion

This research introduces a new role engine designed for multiple collaborative robots. The algorithm focuses on defining, managing, and evaluating the roles of a team of robots to enhance their performance. It incorporates a consensus-based process role optimization that utilizes GP inference, taking into account environmental constraints and shared factors in dynamic multi-robot scenarios. Additionally, an environment map is created based on the field's skeleton to improve feasibility and scalability. Both simulations and real-world experiments involving ground robots of different sizes navigating through obstacles demonstrate the effectiveness and efficiency of the proposed framework.

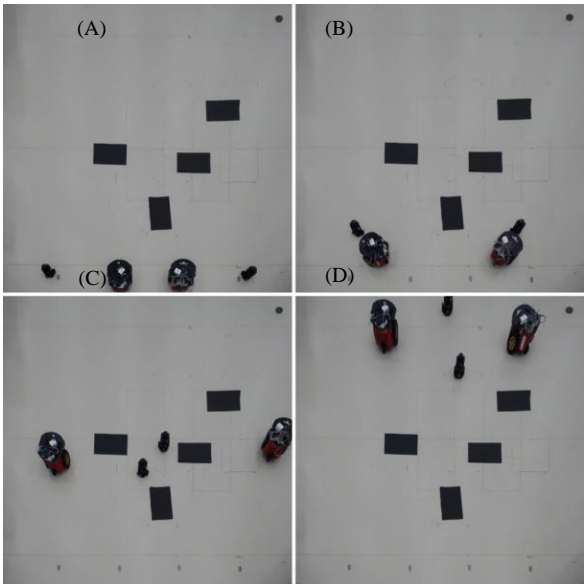

**Fig. 16.** Experimental setup: starting from the positions in (A) the agents move to the diamond formation in (D). In (B), the Turtlebots move in front of the Pioneers, while avoiding collisions, to take efficient paths through the obstacles in (C).

Future work will focus on applying the proposed method to address complex requirements, such as adaptive robot team collaboration using dynamic role assignments. It is important to note that this paper initially employs a central computer for role negotiation, agent evaluation, and role assignment. However, future research will explore the potential of the central computer to play more significant roles in robot team collaboration in an adaptive way. Additionally, there are plans to expand the scale and environment with a fully decentralized approach by developing a consensus algorithm for each robot.

## Appendix A

*A) Environment Map* – An E-MAP is a customized weighted graph that serves as a representation of the environment, specifically designed for a particular type of robot. The process of creating an E-Map involves the following steps:

1. Assign penalties to the obstacles in the environment based on the characteristics of the robot.

2. Generate the skeleton of the environment using a thinning algorithm, [13]. Additionally, remove stair artifacts using a technique similar to [14].

3. Identify feature nodes, which are nodes on the skeleton that have neighbors in more than one direction. These nodes are then used to construct the graph, as depicted in Figure 16.

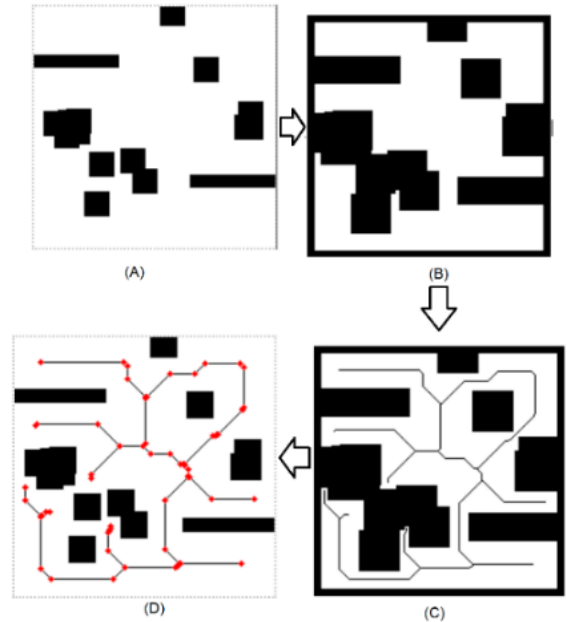

**Fig. 16.** Making E-Map: A) Original Image; B) Calculate the visitable pixels; C) Create the skeleton and de-stairing; D) Make weighted graph from the corner and intersections.

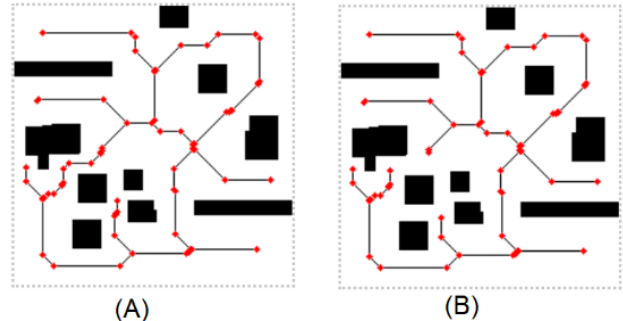

**Fig. 17.** E-Maps for two types of robots: (A) R=0.1; (B) R=0.05

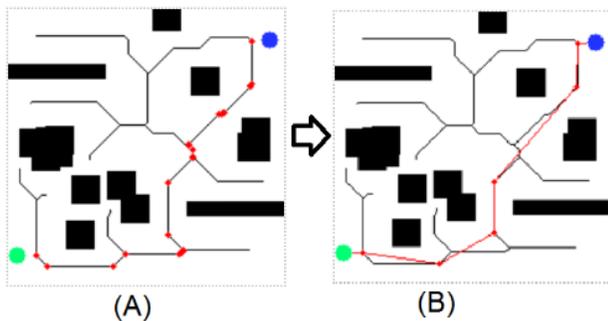

**Fig. 18.** (A) Find the shortest path; (B) Reduce the number of nodes

*B) Detecting Auxiliary nodes* - Auxiliary nodes between two start and end nodes are the nodes in the graph that make the shortest path; the nearest node on the skeleton to the source and destination is considered as the start and end nodes, respectively. Using these nodes and the environment map, A* search algorithm [15] with an octile distance heuristics is applied to find the shortest path on the environment map from the start to the end node, as shown in Fig. 18. Additional filtering may be done to the nodes to reduce the number of nodes. This node reduction is made in consideration of the size of the robot, based on the feasible location map generated earlier. It searches for feasible nodes by drawing a line and checking whether the line passes through obstacles. If no path is found between the start and end node, a path between the source and destination may not be feasible. The feature nodes traversed along the path are returned as the reference auxiliary nodes to speed up the initial path calculation.